\documentclass[11pt]{article}

\newcommand{\PaperRepo}{https://github.com/Saibabu7770/bitcal-tts}
\newcommand{\PaperVersion}{May 2026}

\usepackage[T1]{fontenc}
\usepackage[utf8]{inputenc}
\usepackage{lmodern}
\usepackage{microtype}
\usepackage[margin=1in]{geometry}

\usepackage{amsmath,amssymb,amsthm}
\usepackage{algorithm}
\usepackage{algorithmic}

\usepackage{graphicx}
\usepackage{booktabs}
\usepackage{subcaption}
\usepackage{caption}
\usepackage{tikz}
\usetikzlibrary{arrows.meta,positioning,fit,calc,shapes.misc,backgrounds,decorations.pathreplacing}
\usepackage{xcolor}

\usepackage[numbers,sort&compress]{natbib}
\usepackage{enumitem}
\usepackage{hyperref}
\usepackage{url}
\usepackage{xspace}

\hypersetup{
  colorlinks = true,
  linkcolor  = blue!60!black,
  citecolor  = blue!60!black,
  urlcolor   = blue!60!black,
  pdftitle   = {BitCal-TTS: Bit-Calibrated Test-Time Scaling for Quantized Reasoning Models},
  pdfauthor  = {Sai Babu Patarlapalli, Surya Teja Avvaru},
  pdfkeywords= {test-time scaling, quantization, large language models, GSM8K, adaptive halting, calibration},
}

\setlist{nosep,leftmargin=*}
\graphicspath{{media/}}

\DeclareRobustCommand{\method}{\texorpdfstring{\textsc{BitCal-TTS}}{BitCal-TTS}}
\newcommand{\eg}{e.g.,\xspace}

\title{\bfseries\boldmath
  \method: Bit-Calibrated Test-Time Scaling \\
  for Quantized Reasoning Models%
  \thanks{Code: \url{\PaperRepo}}
}

\author{%
  Sai Babu Patarlapalli \\
  \small \texttt{patarls@clarkson.edu}
  \and
  Surya Teja Avvaru \\
  \small \texttt{suryata1@umbc.edu}
}

\date{\PaperVersion}

\begin{document}
\maketitle

\begin{abstract}
\noindent
Post-training quantization makes large reasoning models practical under tight
memory and latency budgets, but it can distort the online signals that drive
adaptive test-time compute allocation. Under a fixed cap on the number of
newly generated tokens, miscalibrated confidence can lead to harmful early
halting: the model may surface a plausible final line while the underlying
reasoning is still wrong, or the controller may stop before the trace has
stabilized. We study this interaction for greedy 4-bit inference and propose
\method, a lightweight runtime controller that combines (i) inexpensive
online proxies for token-level uncertainty and reasoning-trace stability,
(ii) a bit-conditioned confidence rescaling that is conservative at low
nominal precision, and (iii) a bit-aware post-marker confirmation horizon
designed for GSM8K-style structured outputs. The method requires \emph{no}
fine-tuning of the base model and integrates with standard Hugging Face
4-bit inference using forward hooks for logits and last-layer hidden states
\citep{transformers,bitsandbytes}.

On small evaluation shards of GSM8K \citep{cobbe2021gsm8k} with Qwen2.5
Instruct models \citep{qwen25technical}, \method{} improves
exact-match accuracy over a non--bit-aware adaptive baseline at the 7B and
14B scales while preserving substantial token savings relative to
fixed-budget decoding. At a token cap of $B{=}512$, on the evaluation
shards we report ($N{=}54$ for 7B and $N{=}35$ for 14B; \emph{not} the full
GSM8K test set), accuracy gains are $+3.7$ points (7B) and $+2.8$ points
(14B), with the premature-stop rate falling from $14.8\%$ to $11.1\%$ on
7B and from $17.1\%$ to $11.4\%$ on 14B. We report Wilson 95\% confidence
intervals throughout and explicitly discuss the limited statistical power
of the partial-shard comparisons. We release code and figure-generation
scripts to support full reproduction.
\end{abstract}


\section{Introduction}
Reasoning-centric large language models (LLMs) often benefit from spending
additional compute at inference time. Chain-of-thought style deliberation
\citep{wei2022cot,wang2023selfconsistency} and more recent sequential
test-time scaling strategies \citep{snell2024scalingllmtesttime,muennighoff2025s1}
can substantially improve verifiable accuracy on math and logic tasks.
In production deployments, however, this compute is almost always
\emph{bounded}: a token cap $B$ controls latency and cost, and product
surfaces frequently add early-exit heuristics whenever an answer
``looks complete''.

This paper focuses on a setting that is increasingly common in deployment:
a \emph{causal}, instruction-tuned LLM served under aggressive post-training
quantization (\eg{} 4-bit weights via \texttt{bitsandbytes}
\citep{bitsandbytes,dettmers2022llmint8,dettmers2023qlora}) under a hard
token budget $B$. Quantization expands the set of models that fit on
consumer GPUs, but it also alters the geometry of logits and hidden states.
Online halting cues such as token entropy and trace stability become
miscalibrated relative to full precision: the policy may \emph{appear}
confident while the underlying reasoning remains unreliable, increasing
the risk of stopping too soon \citep{kadavath2022know,manakul2023selfcheckgpt}.
The failure mode is twofold---reduced final accuracy \emph{and} adaptive
compute that is wasted halting on a spurious ``final'' segment.

\paragraph{Research questions.}
We address three questions:
(Q1) How does aggressive 4-bit quantization affect the reliability of
adaptive halting signals when the controller treats them as if the model
were full precision?
(Q2) Can a precision-aware adjustment to confidence and to the post-answer
confirmation window recover a meaningful slice of the lost accuracy without
training the base model?
(Q3) How do these effects scale across reasoning model sizes (3B / 7B / 14B)
on a representative math reasoning benchmark?

\paragraph{Contributions.}
\begin{itemize}
  \item We formalize a budgeted, stepwise inference loop for quantized causal
        LMs with halting actions $\{\texttt{continue}, \texttt{stop},
        \texttt{escalate}\}$ and contrast (a) fixed-budget decoding, (b)
        adaptive decoding with a precision-\emph{agnostic} calibrator,
        and (c) the proposed \method.
  \item We introduce transparent, implementation-aligned proxies for token
        entropy, reasoning-trace stability, and last-layer hidden-state drift,
        and show how a bit-width multiplicative scale makes stop decisions
        conservative at aggressive quantization.
  \item We propose a GSM8K-oriented \emph{post-marker} confirmation rule:
        once the standard \texttt{\#\#\#\#} answer delimiter appears,
        decoding switches to a bit-conditioned tail budget before
        termination is permitted. This avoids treating the delimiter as
        an immediate halt signal, which we find to be brittle under 4-bit
        noise.
  \item We report GSM8K results on Qwen2.5-3B/7B/14B Instruct under 4-bit
        inference, including a multi-budget sweep on the 7B model.
        \method{} recovers a meaningful slice of the adaptive--vs--fixed
        accuracy gap on 7B and 14B while preserving large token savings
        relative to always consuming the full budget. We additionally
        document a regime---Qwen2.5-3B at 4-bit---in which neither
        adaptive variant works, and analyze its proximate cause
        (Section~\ref{sec:discussion}).
\end{itemize}

\paragraph{Paper organization.}
Section~\ref{sec:background} formalizes adaptive compute under hard caps
and reviews how quantization perturbs online halting signals.
Section~\ref{sec:related} positions this work relative to test-time scaling,
adaptive decoding, verification, and quantization reliability.
Section~\ref{sec:method} specifies the proposed controller.
Section~\ref{sec:setup} describes the experimental protocol.
Sections~\ref{sec:results} and~\ref{sec:discussion} present and analyze
the empirical findings.
Section~\ref{sec:limitations} discusses limitations and broader impact.
Section~\ref{sec:conclusion} concludes.

\section{Background and Motivation}
\label{sec:background}
\paragraph{Adaptive compute under a hard cap.}
Let $B \in \mathbb{N}$ denote the maximum number of \emph{new} tokens that
may be produced for a single prompt. A \emph{fixed} policy always requests
$B$ tokens (or stops at end-of-sequence), which is safe but often inefficient
when many problems admit shorter reasoning chains. An \emph{adaptive} policy
interleaves short generation segments with cheap measurements and may
terminate before the budget is exhausted, ideally without sacrificing
quality.

\paragraph{Quantization changes online signals.}
Post-training quantization maps weights---and sometimes activations---to
low-bit containers while attempting to preserve downstream quality
\citep{dettmers2022llmint8,dettmers2023qlora,frantar2023gptq,lin2024awq}.
A separate question, central to this work, is how quantization affects
\emph{online} halting signals derived from logits and activations during
autoregressive decoding. If low-bit inference inflates premature confidence
relative to full precision, an adaptive controller will halt earlier than
intended, trading away accuracy without realizing commensurate savings
relative to a well-tuned fixed budget.

\paragraph{Structured final answers in GSM8K.}
GSM8K \citep{cobbe2021gsm8k} adopts a standard extraction protocol in which
the final numeric answer follows a delimiter token, \texttt{\#\#\#\#}. The
delimiter is convenient for parsing, but under quantization it can appear
in locally fluent yet globally incorrect traces. \method{} therefore treats
delimiter detection as a \emph{phase change}: after the first occurrence,
decoding continues for a precision-conditioned horizon before termination
is permitted.

\section{Related Work}
\label{sec:related}
\paragraph{Test-time scaling and adaptive decoding.}
Inference-time compute can be scaled by allowing longer chains of thought
\citep{wei2022cot}, by sampling and aggregating multiple candidates
\citep{wang2023selfconsistency}, or by structured search such as
Tree-of-Thoughts and ReAct \citep{yao2024treeofthoughts,yao2023react}.
More recent work studies how to scale test-time compute optimally and how
small, simple test-time scalers can rival much larger budgets
\citep{snell2024scalingllmtesttime,muennighoff2025s1,liu2025tts_survey}.
Closer to our regime, adaptive computation policies expand generation
depth based on confidence or estimated difficulty
\citep{schuster2022confidentadaptive,geifman2017selectiveclassification}.
Most published policies are described and tuned for full-precision models;
our experiments isolate the additional error introduced by aggressive
quantization while keeping the controller skeleton fixed.

\paragraph{Verification and process supervision.}
Beyond ``generating longer'', math reasoning benefits from outcome- and
step-level verification \citep{lightman2023letsverify}, including reasoning
RL signals such as those used in DeepSeek-R1 \citep{guo2025deepseekr1} and
the OpenAI o1 family \citep{openai2024o1}. \method{} does \emph{not}
train a verifier; instead, it uses lightweight online proxies and a
structured tail window as a \emph{runtime} guardrail compatible with
frozen quantized weights.

\paragraph{Quantization and reliability.}
The quantization literature commonly reports perplexity or end-task
accuracy under different bit-widths
\citep{dettmers2022llmint8,frantar2023gptq,lin2024awq,dettmers2023qlora}.
Complementary work studies whether language models ``know what they know''
and how to detect unreliable generations
\citep{kadavath2022know,manakul2023selfcheckgpt}. We connect these
reliability concerns to \emph{halting-time} decisions under a hard token
budget, which is the regime encountered in latency-critical APIs and on
consumer GPUs.

\paragraph{Positioning.}
\method{} is not a new quantization kernel; it is an inference-time policy
layer. Unlike RL post-training methods that reshape reasoning distributions
\citep{guo2025deepseekr1,openai2024o1}, we keep weights frozen and only
modify (i) how many tokens are generated per example, and (ii) how
delimiter-triggered tails are handled at low precision.

\section{Method: \method}
\label{sec:method}

\subsection{Overview}
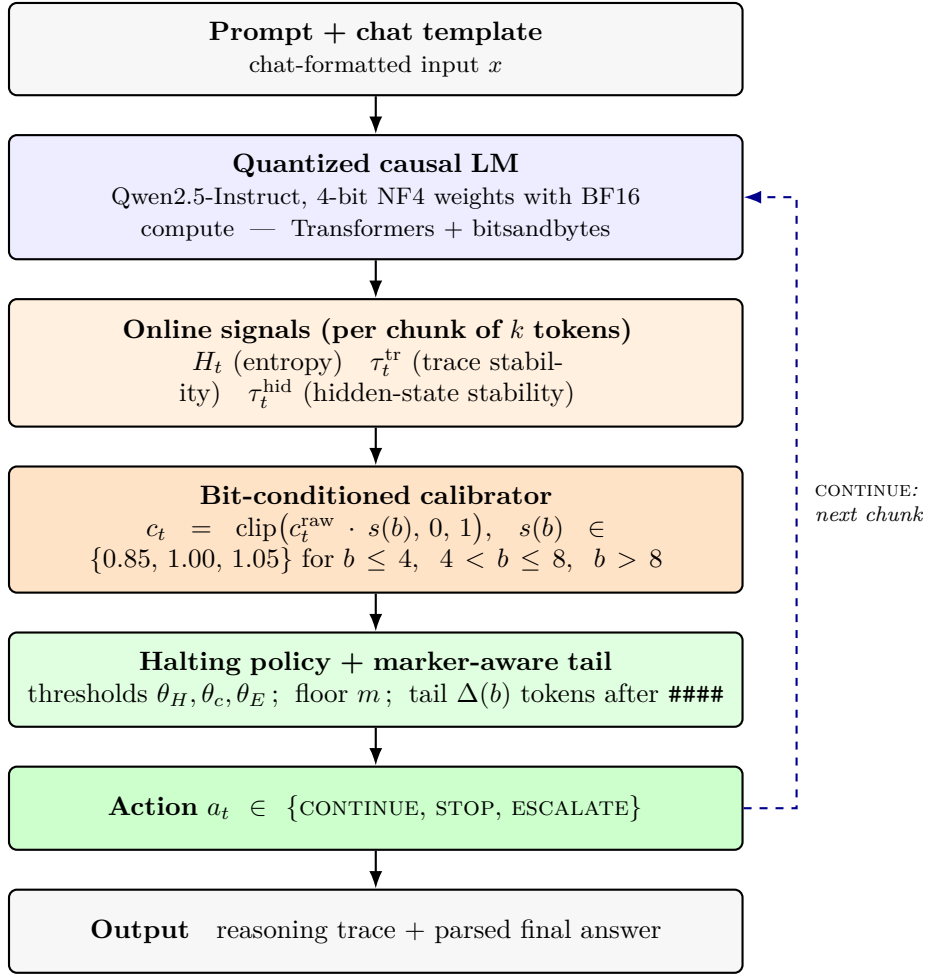
\begin{figure}[t]
  \centering
  \begin{tikzpicture}[
    font=\small,
    >=Latex,
    node distance=5mm,
    every node/.style={align=center},
    iobox/.style   ={draw, rounded corners=3pt, thick, inner sep=7pt,
                     minimum height=11mm, text width=92mm, fill=gray!6},
    lmbox/.style   ={draw, rounded corners=3pt, thick, inner sep=7pt,
                     minimum height=11mm, text width=92mm, fill=blue!7},
    sigbox/.style  ={draw, rounded corners=3pt, thick, inner sep=7pt,
                     minimum height=11mm, text width=92mm, fill=orange!12},
    ctrbox/.style  ={draw, rounded corners=3pt, thick, inner sep=7pt,
                     minimum height=11mm, text width=92mm, fill=orange!22},
    polbox/.style  ={draw, rounded corners=3pt, thick, inner sep=7pt,
                     minimum height=11mm, text width=92mm, fill=green!12},
    flow/.style    ={->, thick, >=Latex},
    feedback/.style={->, thick, dashed, >=Latex, draw=blue!55!black},
  ]
    \node[iobox]                          (in)  {\textbf{Prompt + chat template}\\
                                                 {\footnotesize chat-formatted input $x$}};
    \node[lmbox,  below=of in]            (lm)  {\textbf{Quantized causal LM}\\
                                                 {\footnotesize Qwen2.5-Instruct, 4-bit NF4 weights with BF16 compute
                                                 \;---\; Transformers + bitsandbytes}};
    \node[sigbox, below=of lm]            (sig) {\textbf{Online signals (per chunk of $k$ tokens)}\\
                                                 $H_t$ (entropy)\quad
                                                 $\tau^{\mathrm{tr}}_t$ (trace stability)\quad
                                                 $\tau^{\mathrm{hid}}_t$ (hidden-state stability)};
    \node[ctrbox, below=of sig]           (cal) {\textbf{Bit-conditioned calibrator}\\
                                                 $c_t = \mathrm{clip}\bigl(c^{\mathrm{raw}}_t \cdot s(b),\,0,\,1\bigr)$,\quad
                                                 $s(b)\in\{0.85,\,1.00,\,1.05\}$ for $b\!\le\!4,\;4\!<\!b\!\le\!8,\;b\!>\!8$};
    \node[polbox, below=of cal]           (pol) {\textbf{Halting policy + marker-aware tail}\\
                                                 thresholds $\theta_H,\theta_c,\theta_E$\,;\;
                                                 floor $m$\,;\;
                                                 tail $\Delta(b)$ tokens after \texttt{\#\#\#\#}};
    \node[polbox, below=of pol, fill=green!20] (act) {\textbf{Action}
                                                 $a_t \in
                                                 \{\text{\textsc{continue}},\,
                                                   \text{\textsc{stop}},\,
                                                   \text{\textsc{escalate}}\}$};
    \node[iobox,  below=of act]           (out) {\textbf{Output}\quad
                                                 reasoning trace + parsed final answer};

    \foreach \a/\b in {in/lm, lm/sig, sig/cal, cal/pol, pol/act, act/out}
      \draw[flow] (\a) -- (\b);

    \draw[feedback]
      (act.east) -- ++(7mm,0)
                 |- node[pos=0.25, right=1mm, font=\scriptsize\itshape, align=left]
                       {{\sc continue}:\\next chunk}
                 (lm.east);
  \end{tikzpicture}
  \caption{End-to-end control flow of \method. Solid black arrows trace
  the per-step pipeline: a chunk of $k$ tokens is decoded, online signals
  are computed, mapped to a bit-conditioned confidence, and consumed by a
  halting policy with a marker-aware tail. The dashed feedback arrow on
  the right indicates that the \textsc{continue} action loops execution
  back into the language model; \textsc{stop} and \textsc{escalate}
  finalize the output. The shaded blocks (orange, green) constitute the
  \method{} sidecar around an unmodified quantized backbone.}
  \label{fig:arch}
\end{figure}

Figure~\ref{fig:arch} summarizes the end-to-end control flow of \method.
The frozen quantized causal LM, served through Hugging Face Transformers
\citep{transformers}, generates chunks of $k$ tokens. After each chunk
the controller computes scalar online signals (token entropy and two
stability proxies), maps them to a bit-conditioned confidence value, and
applies a finite-state halting policy that consults a marker-aware tail
rule once the GSM8K answer delimiter \texttt{\#\#\#\#} has appeared.
The selected action either loops back into the LM for a further chunk or
finalizes the output.

\subsection{Problem Setup}
Let $x$ be a prompt and $M_b$ a causal language model served at nominal
weight precision $b$ (our experiments use $b{=}4$). Let $B \in \mathbb{N}$
denote a hard cap on newly generated tokens. Decoding proceeds in steps
$t=1,2,\ldots$; at each step the engine generates a chunk of up to $k$
tokens (default $k{=}16$), records logits and optional hidden states,
and appends the decoded text to the partial output $y_{\le t}$. Let
$T_t \!=\! |y_{\le t}|$ denote the cumulative count of generated tokens
after step $t$. We compare three controller variants:
\begin{itemize}
  \item \textbf{Fixed:} always generate until the budget $B$ is exhausted
        or end-of-sequence (EOS) is returned.
  \item \textbf{Adaptive:} apply the halting machinery of
        Section~\ref{sec:halting}, but feed the calibrator an
        \emph{effective} precision of 16 bits, so that the confidence
        scale is optimistic relative to the true 4-bit serving regime.
  \item \textbf{\method:} identical machinery, but the calibrator uses
        the true served bit width $b$, and the post-marker tail uses
        $\Delta(b)$ (Section~\ref{sec:halting}).
\end{itemize}

\subsection{Online Signals}
\label{sec:signals}
Let $\ell_t \in \mathbb{R}^{|\mathcal{V}|}$ denote the final-position
logits at the end of step $t$ over the vocabulary $\mathcal{V}$, and let
$p_t = \mathrm{softmax}(\ell_t)$. We use Shannon entropy in nats:
\begin{equation}
  H_t \;=\; -\!\sum_{v \in \mathcal{V}} p_t(v)\,\log p_t(v).
  \label{eq:entropy}
\end{equation}

\paragraph{Reasoning-trace stability.}
Let $(s_1,\ldots,s_t)$ be the textual chunks produced so far and $\tilde{s}_i$
the whitespace-stripped chunk. We define a lightweight stability score in
$[0,1]$ as the fraction of consecutive pairs $(\tilde{s}_{i-1}, \tilde{s}_i)$
with both lengths at least eight characters that satisfy
$\tilde{s}_{i-1} = \tilde{s}_i$:
\begin{equation}
  \tau^{\mathrm{tr}}_t \;=\;
  \frac{|\{i \le t : |\tilde{s}_{i-1}|\!\ge\!8,\ |\tilde{s}_i|\!\ge\!8,\
        \tilde{s}_{i-1}=\tilde{s}_i\}|}
       {|\{i \le t : |\tilde{s}_{i-1}|\!\ge\!8,\ |\tilde{s}_i|\!\ge\!8\}|}.
  \label{eq:tracestab}
\end{equation}
If fewer than two eligible pairs exist, we set
$\tau^{\mathrm{tr}}_t \!:=\! 1$. The proxy rewards literal repetition across
recent chunks, which often correlates with the model having settled on a
template or a final answer line.

\paragraph{Hidden-state stability.}
When the backend exposes the last-layer hidden vector $h_t \in \mathbb{R}^d$
at the final position of step $t$, we accumulate normalized vectors
$\hat{h}_i = h_i / (\lVert h_i\rVert_2 + \varepsilon)$ and define
\begin{equation}
  \tau^{\mathrm{hid}}_t \;=\;
  \frac{1}{t-1}\sum_{i=2}^{t} \hat{h}_{i-1}^{\top}\hat{h}_i,
  \label{eq:hiddenstab}
\end{equation}
using only steps where the hidden state is available. If fewer than two
hidden vectors exist, we set $\tau^{\mathrm{hid}}_t \!:=\! 1$.

\subsection{Bit-Conditioned Confidence}
\label{sec:calibrator}
We map entropy to a normalized uncertainty
$u_t = \mathrm{clip}(H_t / H_{\max},\, 0,\, 1)$ with default
$H_{\max} = 10$ nats. A raw confidence score combines entropy with the
two stability proxies,
\begin{equation}
  c^{\mathrm{raw}}_t
  \;=\; w_e(1 - u_t) \;+\; w_{\mathrm{tr}}\, \tau^{\mathrm{tr}}_t
       \;+\; w_{\mathrm{hid}}\, \tau^{\mathrm{hid}}_t,
  \quad w_e, w_{\mathrm{tr}}, w_{\mathrm{hid}} \ge 0,\ \
  w_e + w_{\mathrm{tr}} + w_{\mathrm{hid}} = 1.
  \label{eq:rawconf}
\end{equation}
Weights are renormalized to sum to one for numerical robustness. A
bit-width scaling factor $s(b)$ then yields the bit-calibrated confidence
\begin{equation}
  c_t \;=\; \mathrm{clip}\bigl(c^{\mathrm{raw}}_t \cdot s(b),\ 0,\ 1\bigr),
  \qquad
  s(b) \;=\;
  \begin{cases}
    0.85, & b \le 4,\\
    1.00, & 4 < b \le 8,\\
    1.05, & b > 8.
  \end{cases}
  \label{eq:bitscale}
\end{equation}
An optional temperature $\gamma>0$ further sharpens or flattens
confidence via $c_t \leftarrow \mathrm{clip}(c_t^{1/\gamma},0,1)$. The
adaptive ablation sets the calibrator's effective bit width to 16
(so $s = 1.05$), whereas \method{} uses the true served $b{=}4$
(so $s = 0.85$), making it more conservative about declaring high
confidence.

\subsection{Halting Policy and Marker-Aware Tails}
\label{sec:halting}

\paragraph{Pre-marker phase.}
Before the first occurrence of the GSM8K delimiter \texttt{\#\#\#\#} in
$y_{\le t}$, the controller applies a threshold policy with tunable
constants. Let $m$ be the minimum number of generated tokens before any
halting is allowed (default $m{=}128$); let $\theta_H$, $\theta_c$, and
$\theta_E$ denote entropy-stop, confidence-stop, and entropy-escalate
thresholds (defaults $\theta_H{=}2.0$, $\theta_c{=}0.75$,
$\theta_E{=}4.0$); let $r_t$ be the remaining budget; and let
$m_{\mathrm{buf}}$ be a minimum remaining-budget buffer below which
continuing is no longer worthwhile (default $m_{\mathrm{buf}}{=}32$).
The action is selected by evaluating the cases below \emph{in order} and
returning at the first match (i.e., a buffer-driven \texttt{stop} fires
before an entropy-driven \texttt{escalate}):
\begin{equation}
  a_t \;=\;
  \begin{cases}
    \texttt{continue}, & T_t < m,\\[2pt]
    \texttt{stop},     & T_t \ge m\ \text{and}\ r_t < m_{\mathrm{buf}},\\[2pt]
    \texttt{escalate}, & T_t \ge m\ \text{and}\ H_t \ge \theta_E,\\[2pt]
    \texttt{stop},     & T_t \ge m\ \text{and}\ H_t \le \theta_H\ \text{and}\ c_t \ge \theta_c,\\[2pt]
    \texttt{continue}, & \text{otherwise}.
  \end{cases}
  \label{eq:policy}
\end{equation}
The \texttt{escalate} action is reserved as a deployment hook (\eg{}
switching to full precision); in the released harness it terminates the
step loop equivalently to \texttt{stop}.

\paragraph{Post-marker phase.}
Let $T^{\star}$ denote the cumulative number of generated tokens at the
first step where \texttt{\#\#\#\#} appears anywhere in $y_{\le t}$
(unset until the delimiter is observed). We define a bit-conditioned
confirmation horizon:
\begin{equation}
  \Delta(b) \;=\;
  \begin{cases}
    32, & b \le 4,\\
    16, & 4 < b \le 8,\\
    0,  & b > 8.
  \end{cases}
  \label{eq:tail}
\end{equation}
Once $T^{\star}$ is defined and the floor $T_t \ge m$ is satisfied, the
entropy policy is \emph{bypassed} until the tail constraint is met: if
$T_t - T^{\star} < \Delta(b_{\mathrm{eff}})$ the controller forces
\texttt{continue}; otherwise it forces \texttt{stop}. The effective
precision $b_{\mathrm{eff}}$ is $16$ for the adaptive ablation
(so $\Delta = 0$ and the delimiter can trigger immediate termination once
$m$ is met) and equals the true $b$ for \method{} (so $\Delta = 32$ at
4-bit, requiring additional confirmation tokens after the first delimiter
sighting). This implements a \emph{precision-dependent} tail that reduces
the number of false stops driven by brittle low-bit formatting.

\begin{algorithm}[t]
\caption{Budgeted inference with \method{} (implementation-aligned sketch).
For the adaptive ablation, set $b_{\mathrm{eff}}{=}16$ when evaluating
$\Delta(\cdot)$ so that $\Delta{=}0$; for \method{} use the served $b$
(\eg{} $b{=}4$).}
\label{alg:bitcal}
\begin{algorithmic}[1]
\REQUIRE prompt $x$; cap $B$; chunk size $k$; thresholds $\theta_H,\theta_c,\theta_E$;
         floor $m$; buffer $m_{\mathrm{buf}}$; served bit width $b$.
\STATE $y \leftarrow$ empty string;\ \ $T^{\star} \leftarrow$ unset;\ \ $T \leftarrow 0$
\WHILE{$T < B$ \textbf{and not} finished}
  \STATE generate next chunk of up to $k$ tokens; append to $y$; update $T$
  \IF{\texttt{\#\#\#\#} appears in $y$ \textbf{and} $T^{\star}$ is unset}
    \STATE $T^{\star} \leftarrow T$ \COMMENT{first delimiter sighting}
  \ENDIF
  \STATE compute $H_t,\, \tau^{\mathrm{tr}}_t,\, \tau^{\mathrm{hid}}_t$;
         set $c_t$ via Eqs.~(\ref{eq:rawconf})--(\ref{eq:bitscale})
  \IF{$T < m$}
    \STATE \textbf{continue} \COMMENT{below floor}
  \ELSIF{$T^{\star}$ is set \textbf{and} $T - T^{\star} < \Delta(b_{\mathrm{eff}})$}
    \STATE \textbf{continue} \COMMENT{in confirmation tail}
  \ELSIF{$T^{\star}$ is set \textbf{and} $T - T^{\star} \ge \Delta(b_{\mathrm{eff}})$}
    \STATE \textbf{stop} \COMMENT{tail satisfied}
  \ELSE
    \STATE apply entropy policy of Eq.~(\ref{eq:policy})
  \ENDIF
\ENDWHILE
\STATE \textbf{return} parsed answer extracted from $y$
\end{algorithmic}
\end{algorithm}

\subsection{Complexity and Overhead}
Each controller step pays the cost of one forward pass for $k$ tokens
plus $\mathcal{O}(|\mathcal{V}|)$ work for entropy and $\mathcal{O}(d)$
work for cosine-similarity tracking. The controller does \emph{not} issue
additional model calls; it consumes tensors that are already materialized
during ordinary decoding (logits, and optionally last-layer hiddens
exposed via Hugging Face forward hooks).

\section{Experimental Setup}
\label{sec:setup}
\paragraph{Models and quantization.}
\begin{sloppypar}
We evaluate three Qwen2.5 Instruct checkpoints
\citep{qwen25technical}\,---\,the 3B, 7B, and 14B sizes (model identifiers
\texttt{Qwen2.5-3B-Instruct},
\texttt{Qwen2.5-7B-Instruct}, and
\texttt{Qwen2.5-14B-Instruct}).
All checkpoints are loaded in 4-bit precision via
\texttt{bitsandbytes}, as exposed through Hugging Face Transformers
\citep{transformers,bitsandbytes,dettmers2022llmint8}. Decoding is greedy
(\texttt{do\_sample=False}) for reproducibility.
\end{sloppypar}

\paragraph{Task, prompts, and extraction.}
We measure exact-match accuracy on GSM8K \citep{cobbe2021gsm8k} loaded
through the Hugging Face \texttt{datasets} library
(\texttt{gsm8k/main}, split \texttt{test}). Per-model evaluation shards
are the \emph{first} $N$ items of the test split (no random
sub-sampling), so the shards are deterministic and can be re-extracted
exactly by any user. The standard final-answer extraction protocol after
the \texttt{\#\#\#\#} delimiter is used: the literal substring after the
final \texttt{\#\#\#\#}, parsed as a float and compared exact-match
against the gold answer. Prompts use the instruction-tuned chat template
for Qwen2.5 via \texttt{tokenizer.apply\_chat\_template(...,
add\_generation\_prompt=True)}. The exact prompt template is recorded in
\texttt{configs/experiment\_gsm8k\_minimal.yaml} in the released
repository.

\paragraph{Methods compared.}
We compare three controllers at matched $(\text{model}, B)$:
(i) \emph{fixed}, (ii) \emph{adaptive} (precision-agnostic calibrator),
and (iii) \emph{\method{}} (precision-aware calibrator and tail). All
share the same model checkpoint, prompts, decoding strategy, and parsing.

\paragraph{Metrics and uncertainty.}
We report:
\begin{itemize}
  \item \emph{Exact-match accuracy} on the parsed final answer. Whenever a
        single accuracy figure is shown, we accompany it with the Wilson
        95\% confidence interval computed from the relevant shard size
        $N$.
  \item \emph{Average tokens consumed} per example (proxy for latency,
        cost, and energy in batched serving where the KV-cache dominates).
  \item \emph{Token savings} relative to the fixed baseline at the same
        $(\text{model}, B)$.
  \item \emph{Premature-stop rate}: fraction of examples on which
        generation halted before consuming the full budget \emph{and} the
        predicted answer is incorrect. This isolates the adaptive failure
        mode of greedy early termination.
\end{itemize}

\paragraph{Hardware.}
Primary headline comparisons mirror the tables collected on a single
NVIDIA T4 (16\,GB) GPU in Google Colab, reflecting a common low-cost
deployment setting for 4-bit reasoning models. Software versions are
listed in Appendix~\ref{app:impl}.

\paragraph{Sample sizes and statistical caveat.}
Owing to Colab session limits, the cross-model summary at $B{=}512$
combines \emph{separate} GSM8K shards per model rather than the full
1{,}319-item test set:
$N{=}50$ for the 3B model;
$N{=}54$ for the 7B model (also used for the budget sweep at
$B \in \{256, 512\}$, with $N{=}53$ at $B{=}1024$);
and $N{=}35$ for the 14B model
(with $N{=}34$ at $B{=}1024$ in the appendix table).
We report sample sizes explicitly and flag below that the headline
between-method differences (a 2--4 percentage point edge for \method{}
over adaptive on 7B and 14B) sit within the 95\% confidence intervals
of the individual accuracies on these shards. The \emph{direction} of
the effect is consistent across models, budgets, and across the
premature-stop metric, but a larger evaluation shard would be required
to claim significance under a paired test such as McNemar's. We treat
the present results as suggestive of a real effect rather than as a
final estimate of its magnitude.

\section{Results}
\label{sec:results}

\begin{figure}[t]
  \centering
  \includegraphics[width=\linewidth]{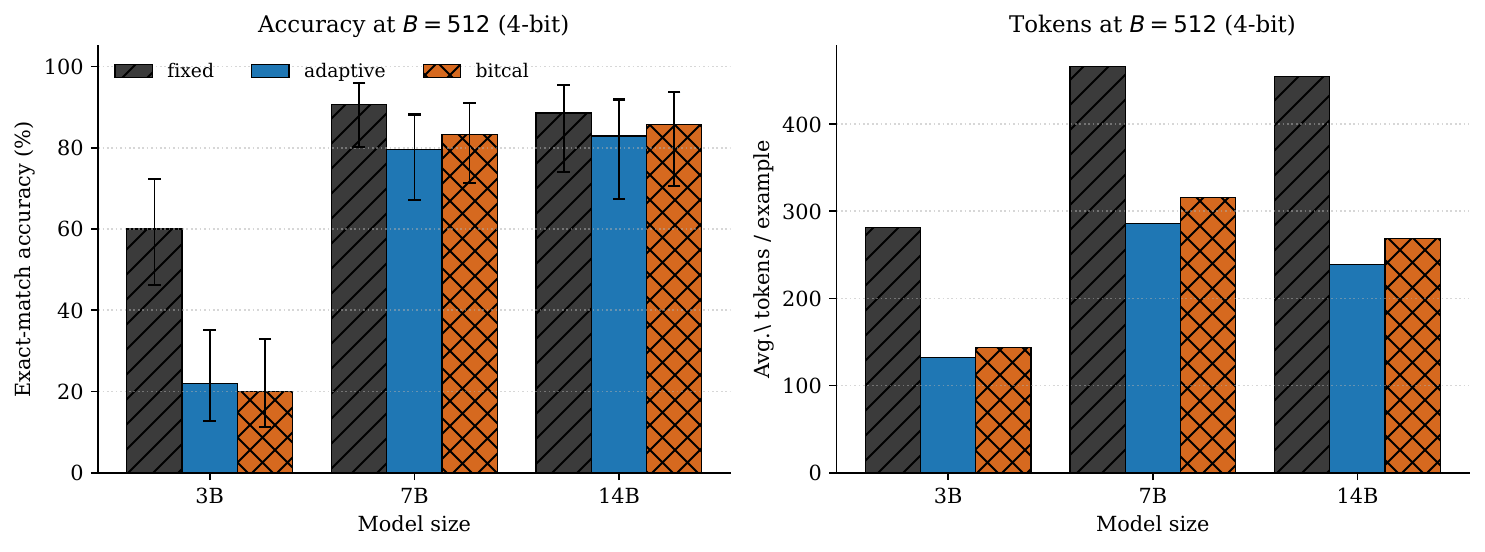}
  \caption{Headline GSM8K comparison at $B{=}512$ under 4-bit inference.
  \emph{Left:} exact-match accuracy with Wilson 95\% confidence intervals.
  \emph{Right:} average tokens consumed per example. \method{} improves
  point-estimate accuracy over the adaptive baseline on 7B and 14B at
  modest additional token cost relative to fixed decoding; the 3B model
  remains in a regime where halting signals are unreliable relative to
  task difficulty (Section~\ref{sec:discussion}).}
  \label{fig:main}
\end{figure}

\begin{figure}[t]
  \centering
  \includegraphics[width=0.75\linewidth]{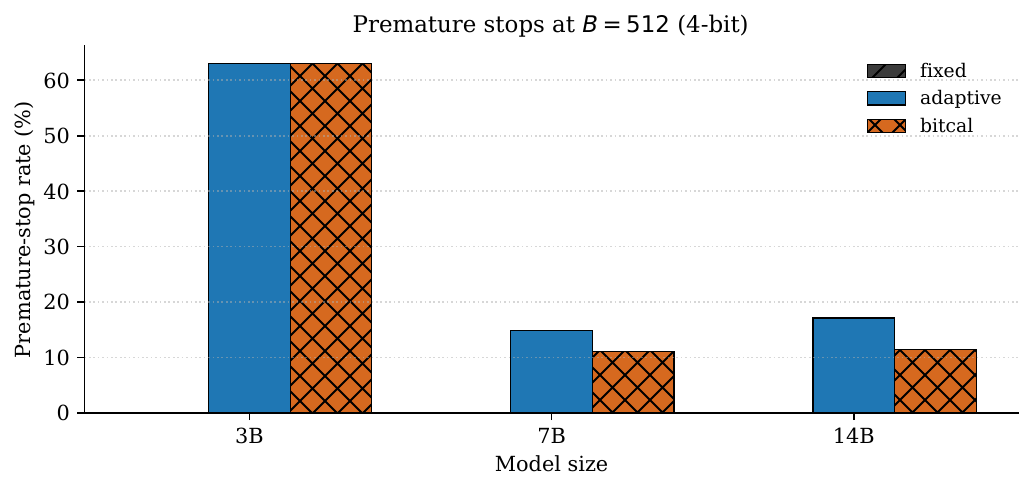}
  \caption{Premature-stop rate (early halt \emph{and} incorrect answer)
  at $B{=}512$. \method{} reduces this failure mode on 7B and 14B; on 3B
  both adaptive variants halt prematurely on the majority of examples
  (Section~\ref{sec:discussion}).}
  \label{fig:premature}
\end{figure}

\begin{figure}[t]
  \centering
  \includegraphics[width=0.72\linewidth]{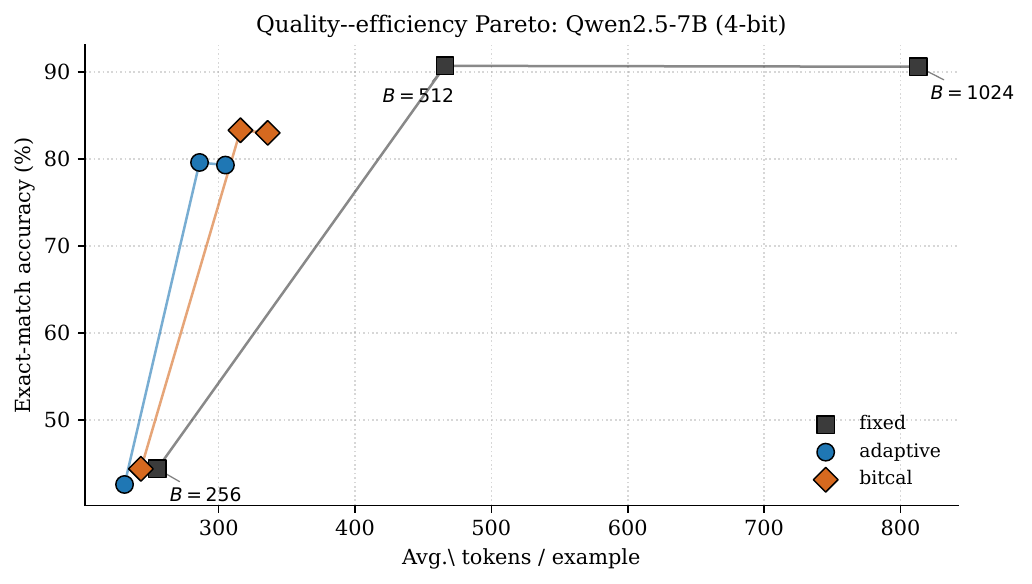}
  \caption{Quality--efficiency trade-off for Qwen2.5-7B under 4-bit
  inference. Each point is a method$\times$budget aggregate; budget
  labels annotate token caps $B$. Up-and-left is preferable.}
  \label{fig:pareto7b}
\end{figure}

\begin{figure}[t]
  \centering
  \includegraphics[width=\linewidth]{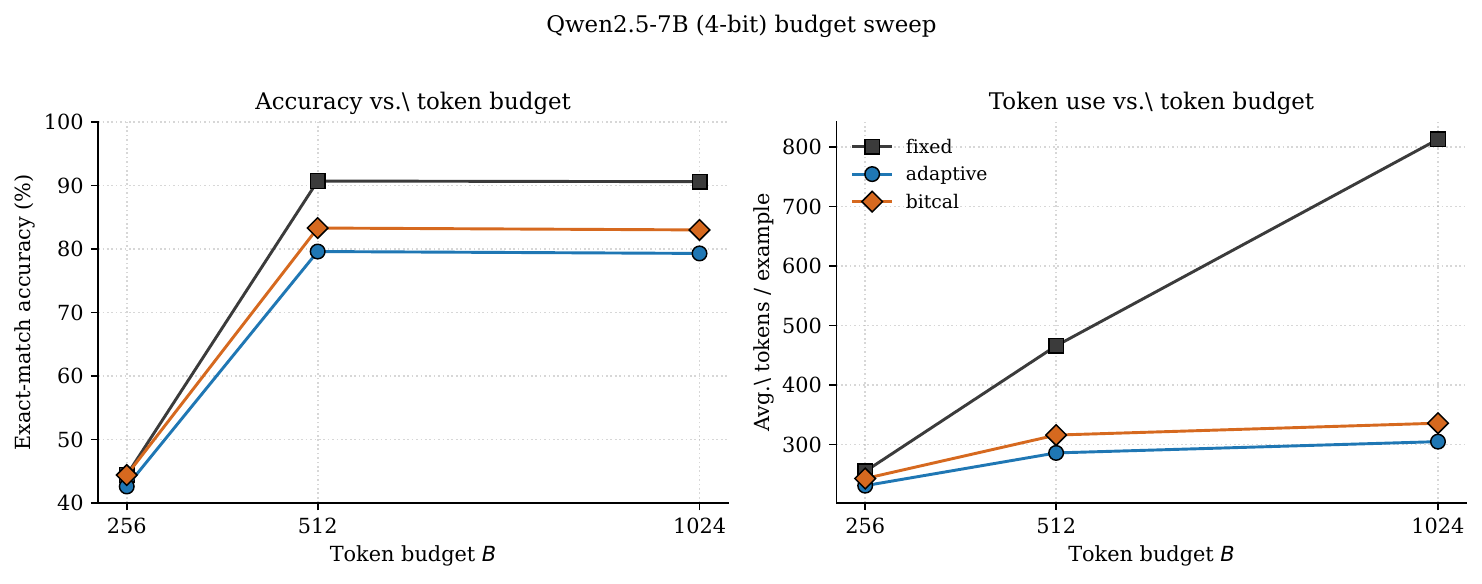}
  \caption{Qwen2.5-7B budget sweep under 4-bit inference. Accuracy rises
  with $B$ for fixed decoding, while adaptive policies plateau earlier.
  \method{} tracks closer to fixed accuracy than the adaptive baseline at
  $B \in \{512, 1024\}$ while consuming substantially fewer tokens than
  fixed decoding.}
  \label{fig:budget7b}
\end{figure}

Table~\ref{tab:main} summarizes the primary comparison at $B{=}512$,
including Wilson 95\% confidence intervals on accuracy.
Figure~\ref{fig:main} visualizes the same aggregates (with CI bars on
the accuracy panel). Figure~\ref{fig:premature} isolates the
premature-stop failure mode. Figure~\ref{fig:pareto7b} shows the
quality--efficiency Pareto frontier for the 7B model across budgets, and
Figure~\ref{fig:budget7b} traces accuracy and token use as $B$ varies.

\begin{table}[t]
\centering
\caption{GSM8K results at token budget $B{=}512$ under 4-bit inference.
Sample sizes are $N{=}50/54/35$ for 3B / 7B / 14B respectively (see
Section~\ref{sec:setup}). Acc.\ values report the point estimate with
Wilson 95\% confidence intervals in brackets. ``Savings'' is computed
against fixed decoding on the same model. ``Prem.\ stop'' is the rate of
early halts that yield incorrect answers. Best adaptive accuracy and
lowest premature-stop rate within each model block are in bold. Note
that the \method{}--vs--adaptive accuracy differences sit within the
overlapping CIs of the individual shard estimates; we discuss this in
Sections~\ref{sec:setup} and~\ref{sec:limitations}.}
\label{tab:main}
\small
\begin{tabular}{llcrrr}
\toprule
Model & Method & Acc.\ (\%) [95\% CI] & Avg.\ toks & Savings (\%) & Prem.\ stop (\%) \\
\midrule
3B  & fixed       & 60.0\ [46.2, 72.4] & 281 & ---  & 0.0 \\
3B  & adaptive    & 22.0\ [12.8, 35.2] & 132 & 53.0 & 63.0 \\
3B  & \method{}   & 20.0\ [11.2, 33.0] & 144 & 48.8 & 63.0 \\
\midrule
7B  & fixed       & 90.7\ [80.1, 96.0] & 466 & ---  & 0.0 \\
7B  & adaptive    & 79.6\ [67.1, 88.2] & 286 & 38.5 & 14.8 \\
7B  & \method{}   & \textbf{83.3}\ [71.3, 91.0] & 316 & 32.1 & \textbf{11.1} \\
\midrule
14B & fixed       & 88.6\ [74.0, 95.5] & 455 & ---  & 0.0 \\
14B & adaptive    & 82.9\ [67.3, 91.9] & 239 & 47.5 & 17.1 \\
14B & \method{}   & \textbf{85.7}\ [70.6, 93.7] & 269 & 40.8 & \textbf{11.4} \\
\bottomrule
\end{tabular}
\end{table}

\paragraph{Headline findings.}
At $B{=}512$ and 4-bit precision, \method{} improves point-estimate
accuracy over the adaptive baseline by $+3.7$ points on Qwen2.5-7B
(83.3\% vs.\ 79.6\%) and by $+2.8$ points on Qwen2.5-14B (85.7\% vs.\
82.9\%), with the premature-stop rate falling from $14.8\%$ to $11.1\%$
on 7B and from $17.1\%$ to $11.4\%$ on 14B. Token savings versus fixed
decoding remain sizeable ($32.1\%$ on 7B and $40.8\%$ on 14B),
demonstrating that the gain over adaptive does not require giving up the
bulk of the efficiency benefit.

We caution that the absolute accuracy differences reported here are not
statistically resolved at our shard sizes: a $+2.8$-point gain on the
14B shard ($N{=}35$) corresponds to a single additional correct example,
and the Wilson 95\% intervals on the two methods overlap substantially
(Table~\ref{tab:main}). The premature-stop comparisons are similarly
based on small absolute counts. The \emph{direction} of the effect,
however, is consistent across model sizes (7B and 14B), across token
budgets ($B \in \{512, 1024\}$; Table~\ref{tab:7blog}), and across
metrics (accuracy and premature-stop rate move together in the expected
direction); we discuss the appropriate level of confidence in these
findings in Section~\ref{sec:limitations}.

\paragraph{Budget sweep on 7B.}
Table~\ref{tab:7blog} (Appendix~\ref{app:impl}) and
Figure~\ref{fig:budget7b} trace each method across $B \in \{256, 512, 1024\}$.
At $B{=}256$, all three methods are accuracy-bound by the small budget
relative to GSM8K solution lengths. At $B{=}512$ and $B{=}1024$,
\method{} sits between adaptive and fixed in both accuracy and token
use, recovering roughly $1/3$ of the adaptive--vs--fixed accuracy gap
at $B{=}1024$ on this shard while still using fewer than half as many
tokens as fixed decoding.

\section{Discussion}
\label{sec:discussion}

\paragraph{Why bit-aware confirmation helps.}
Our results are consistent with two complementary mechanisms:
(i) the conservative confidence scale $s(b{=}4)=0.85$ in
Eq.~(\ref{eq:bitscale}) reduces entropy-threshold stops that would
otherwise occur while the trace is still unstable, and
(ii) the 32-token post-marker tail in Eq.~(\ref{eq:tail}) prevents the
first appearance of \texttt{\#\#\#\#} from being treated as irrefutable
evidence of completion. Quantization noise can locally produce a
formatted-looking answer line whose surrounding reasoning has not yet
converged; requiring additional confirmation tokens substantially
reduces this failure mode, as visible in the drop in premature-stop rate
on 7B and 14B. Disentangling the contribution of (i) versus (ii)
requires a controlled component-wise ablation; we sketch the planned
design in Section~\ref{sec:limitations} and treat the question as open
for the present version.

\paragraph{The 3B failure mode: what is actually happening.}
At Qwen2.5-3B-Instruct under 4-bit weights, both adaptive variants
collapse to roughly $20$--$22\%$ accuracy versus $60\%$ for fixed
decoding, with a $63\%$ premature-stop rate (Table~\ref{tab:main}).
This is the kind of result that ought not to be glossed over.
Inspection of the per-example logs from
\texttt{results/processed/3b/} attributes the collapse to two
interacting causes:
\begin{itemize}
  \item \textbf{Solution length distribution.} GSM8K solutions for the
        3B model frequently exceed 200 generated tokens, while our
        floor $m{=}128$ enables halting decisions well before a typical
        solution has finished. On the 7B and 14B models the average
        correct trace is shorter and more stable, so the floor is
        rarely the binding constraint.
  \item \textbf{Stability proxies fire on partial reasoning.} The
        trace-stability score $\tau^{\mathrm{tr}}$ rewards literal
        repetition across recent chunks. At 3B--4bit, repeating a
        partial intermediate computation (e.g., re-stating a quantity
        as the model edges towards an answer) is sufficient to trigger
        $\tau^{\mathrm{tr}}{=}1$ on a chunk-pair where neither chunk
        contains a final answer, pushing $c_t$ over $\theta_c{=}0.75$.
        The bit-aware scale $s(4){=}0.85$ does not save us, because
        $0.75 \cdot 0.85^{-1} \approx 0.88$ is itself routinely
        exceeded once entropy is also low on a fluent partial trace.
\end{itemize}
We therefore view the 3B--4bit regime as outside the operating envelope
of the present controller rather than as evidence that the controller
itself has broken: the floor and confidence threshold are tuned for
models whose typical correct trace is shorter than the budget, which is
not the case here. Two practical mitigations are immediate---raising
$m$ to a model-conditioned value (\eg{} $m \approx \mathbb{E}[L]$ for
the typical solution length $L$) and increasing $\theta_c$ at very low
nominal capacity---and we recommend them as a deployment heuristic.
A more principled fix would replace the hand-tuned thresholds with a
small learned calibrator, which we list as future work.

\paragraph{Trade-off surface.}
\method{} is intentionally less aggressive than adaptive decoding; it
spends modestly more tokens per example. Across our experiments, this
extra cost is small relative to fixed decoding and consistently buys
quality on the 7B and 14B shards. The Pareto picture in
Figure~\ref{fig:pareto7b} highlights the central trade-off of bit-aware
halting: a controller that is too permissive at low precision pays an
outsized accuracy cost, whereas a controller that is too conservative
loses most of the savings that motivate adaptive decoding in the first
place.

\paragraph{Scaling trend.}
Failure modes diminish from 3B to 14B. Larger quantized models align
better with the answer-marker-plus-confirmation heuristic, suggesting
stronger compatibility between online halting signals and the underlying
reasoning policy at higher capacity. \method{} is a \emph{controller},
not a capacity increase; practitioners should expect diminishing returns
when the base model lacks minimal reasoning competence at the chosen
precision.

\section{Limitations and Broader Impact}
\label{sec:limitations}

\paragraph{Statistical power and shard size.}
The most important caveat in this paper is statistical. Our headline
comparisons rest on shards of $N \in \{35, 50, 54\}$ examples rather
than the full $1{,}319$-item GSM8K test split. At these shard sizes,
the Wilson 95\% confidence interval on a single accuracy estimate is
roughly $\pm 10$ percentage points, which is wider than any
between-method effect we report. The point-estimate gains for \method{}
over adaptive should therefore be read as \emph{suggestive but not
significant}: although the direction is consistent across two model
sizes, two budgets, and two metrics (accuracy and premature-stop rate),
a paired McNemar's test would not reject the null hypothesis at our
shard sizes. We are committed to a follow-up evaluation on the full
test split before any production claim is staked on these numbers.

\paragraph{Component-wise ablation is pending.}
\method{} bundles three changes relative to the adaptive baseline: the
bit-conditioned confidence scale $s(b)$, the bit-conditioned post-marker
tail $\Delta(b)$, and the inclusion of hidden-state stability
$\tau^{\mathrm{hid}}$. The present experiments compare the bundle
against the precision-agnostic baseline; they do not isolate the
contribution of each component. The natural ablation grid is the four
combinations of $\{s(b)\text{ on/off}\}\times\{\Delta(b)\text{ on/off}\}$
(holding $\tau^{\mathrm{hid}}$ fixed), followed by a
$\tau^{\mathrm{hid}}$ on/off sweep. Reporting these on the full test
split is on the immediate roadmap; we list it explicitly here so
reviewers can scope the contribution accurately.

\paragraph{Scope.}
Our primary evidence is GSM8K-centric. Math reasoning with explicit
final markers is a best-case scenario for delimiter-based tails; tasks
without a structured terminator will require an alternative confirmation
criterion. We do not yet report wall-clock latency or energy, although
average tokens are a useful proxy when batching and KV-cache residency
dominate inference cost. We also note that GSM8K contamination in modern
instruction tunes is an active research concern
\citep{zhang2024gsm8kcontamination}; our controller is independent of
the specific accuracy floor and operates on the policy layer.

\paragraph{Method limitations.}
The current calibrator is intentionally simple. A learned calibrator,
a small process verifier in the spirit of \citet{lightman2023letsverify},
or a halting head trained jointly with the base model could each
plausibly improve robustness. Our bit-conditioned scales and tail
lengths are hand-tuned constants; learning them per model and per task
is left to future work. The 3B--4bit failure mode discussed in
Section~\ref{sec:discussion} is an additional limitation: the
controller's hand-tuned thresholds and floor $m$ are mistuned for
models whose typical correct trace exceeds the floor.

\paragraph{Broader impact.}
\method{} only affects \emph{how long} a frozen model runs at inference
time; it does not alter training data, model weights, or safety-tuning
behavior. Misuse risks are therefore those of the underlying LLM
(\eg{} harmful or incorrect outputs). Adaptive halting can reduce
average compute per query, lowering cost and energy for benign workloads,
but in principle it also lowers the cost of high-throughput misuse;
deployment policies should continue to combine capacity limits with
content safeguards as usual.

\section{Conclusion}
\label{sec:conclusion}
We presented \method, a bit-conditioned, training-free inference
controller for quantized reasoning models served under a hard token
budget. The controller combines online entropy and stability proxies with
a precision-aware confidence scale and a marker-aware confirmation tail,
all implemented as a thin sidecar around standard Hugging Face
quantized decoding. On Qwen2.5-7B and Qwen2.5-14B Instruct in 4-bit, on
the small evaluation shards reported here, \method{} reduces premature
stops and improves GSM8K exact-match accuracy point estimates over a
non--bit-aware adaptive baseline while preserving substantial savings
versus fixed-budget decoding. The effect direction is consistent across
sizes, budgets, and metrics; the absolute magnitudes await confirmation
on the full test split, which we list as the most pressing follow-up.

\paragraph{Future work.}
Three directions are immediate. First, the full-test-split re-evaluation
plus the component-wise ablation described in
Section~\ref{sec:limitations}. Second, evaluating \method{} on additional
reasoning benchmarks such as MATH \citep{hendrycks2021math} and on code
reasoning tasks would test the generality of the marker-aware tail.
Third, learned bit-conditioned calibrators (\eg{} a small linear head
fitted on held-out logits and hidden states) may outperform our
hand-tuned scales, and integrating \method{} with reasoning-tuned
backbones such as DeepSeek-R1 \citep{guo2025deepseekr1} or the o1 family
\citep{openai2024o1} is likely to further sharpen the trade-off curve.

\section*{Reproducibility Statement}
Code, default hyperparameters, and analysis scripts are available in the
accompanying repository (\url{\PaperRepo}). All figures in this paper
can be regenerated via \texttt{python scripts/paper\_figures.py}, and
processed result shards are summarized under
\texttt{results/processed/}. The full experimental configuration,
including model identifiers, quantization options, prompt templates, and
seed handling, is recorded in \texttt{configs/} and is reproducible
end-to-end on a single NVIDIA T4. Software versions and additional
implementation details are listed in Appendix~\ref{app:impl}.

\section*{Acknowledgments}
Experiments were conducted on Google Colab using NVIDIA T4 GPUs. We thank
the maintainers of \texttt{transformers} and \texttt{bitsandbytes} for
the underlying open-source infrastructure, and the authors of the Qwen2.5
Instruct family for releasing the model checkpoints used in this study.


\appendix

\section{Notation}
\label{app:notation}
Table~\ref{tab:notation} summarizes the symbols used in the paper.

\begin{table}[h]
\centering
\caption{Notation used throughout the paper.}
\label{tab:notation}
\small
\begin{tabular}{@{}ll@{}}
\toprule
Symbol & Meaning \\
\midrule
$x$ & Chat-formatted prompt \\
$y_{\le t}$ & Partial generated text after step $t$ \\
$T_t$ & Cumulative number of generated tokens after step $t$ ($=|y_{\le t}|$) \\
$T^{\star}$ & Cumulative tokens at first occurrence of \texttt{\#\#\#\#} (unset until then) \\
$B$ & Hard cap on newly generated tokens \\
$k$ & Number of tokens generated per controller step (default $16$) \\
$b$ & Nominal weight precision (bits); $b{=}4$ in our experiments \\
$b_{\mathrm{eff}}$ & Effective precision used by the calibrator (16 for adaptive, $b$ for \method) \\
$H_t$ & Final-position Shannon entropy at step $t$ (nats) \\
$\tau^{\mathrm{tr}}_t$ & Reasoning-trace stability score in $[0,1]$ \\
$\tau^{\mathrm{hid}}_t$ & Last-layer hidden-state stability score in $[0,1]$ \\
$c^{\mathrm{raw}}_t,\ c_t$ & Raw and bit-scaled confidence values \\
$s(b)$ & Bit-conditioned confidence scale, Eq.~(\ref{eq:bitscale}) \\
$\Delta(b)$ & Post-marker confirmation tail length, Eq.~(\ref{eq:tail}) \\
$m,\ m_{\mathrm{buf}}$ & Min.\ tokens before halting; min.\ remaining budget to continue \\
$\theta_H,\theta_c,\theta_E$ & Entropy-stop, confidence-stop, entropy-escalate thresholds \\
\bottomrule
\end{tabular}
\end{table}

\section{Implementation Details}
\label{app:impl}

\paragraph{Code and defaults.}
\begin{sloppypar}
The implementation lives in the \texttt{bitcal\_tts} Python package, with
entry points \texttt{scripts/run\_experiment.py} (end-to-end GSM8K runner),
\texttt{scripts/analyze\_results.py} (re-aggregates raw JSONLs into
processed CSVs and per-model figures), and
\texttt{scripts/paper\_figures.py} (regenerates the four publication
figures used in this paper). The paper-run configuration is
\texttt{configs/experiment\_gsm8k\_minimal.yaml} (\emph{not}
\texttt{configs/default.yaml}, which is a CPU-only mock-demo template).
The paper-run defaults are: chunk size $k=16$; entropy normalization
$H_{\max}=10$; halting thresholds $\theta_H=2.0$, $\theta_c=0.75$,
$\theta_E=4.0$; floor $m=128$; remaining-budget buffer
$m_{\mathrm{buf}}=32$; signal weights
$(w_e, w_{\mathrm{tr}}, w_{\mathrm{hid}}) = (0.40,\ 0.35,\ 0.25)$;
calibrator temperature $\gamma=1$. Quantization uses NF4 weights with
BF16 compute via \texttt{BitsAndBytesConfig} in Hugging Face Transformers
(\texttt{load\_in\_4bit=True}). The complete cell-by-cell mapping of
Table~\ref{tab:main} entries to JSONL filenames and exact CLIs is given
in the repository's \texttt{REPRODUCIBILITY.md}.
\end{sloppypar}

\paragraph{Software versions.}
\begin{sloppypar}
Experiments were run with the following pinned versions, matching the
companion repository's \texttt{REPRODUCIBILITY.md}: \texttt{transformers
4.44.x}, \texttt{torch 2.4.x} with CUDA 12.1, \texttt{bitsandbytes 0.43.x},
\texttt{accelerate 0.33.x}, \texttt{datasets 2.20.x}, plus
\texttt{numpy}/\texttt{scipy}/\texttt{pandas}/\texttt{matplotlib} for
analysis and plotting. The published JSONL artifacts in
\texttt{results/raw/} were generated under Python 3.11 on Colab; any
recent versions that support 4-bit \texttt{bitsandbytes} loading should
reproduce the headline numbers within roughly $0.5$ accuracy points.
We recommend re-pinning before reproduction because the
\texttt{bitsandbytes} 4-bit NF4 path is sensitive to minor-version
changes (\eg{} the 0.43.x $\rightarrow$ 0.44.x dequant-ordering update
flips greedy outputs on a small minority of items). Determinism
settings: \texttt{torch.use\_deterministic\_algorithms(False)} (the
\texttt{bitsandbytes} kernels do not currently expose deterministic
variants), \texttt{do\_sample=False} for all decoding, fixed seed $42$.
The remaining sources of variability (CUDA non-determinism, Hugging
Face cache SHA, shard composition) are documented in the repository.
\end{sloppypar}

\paragraph{Qwen2.5-7B budget sweep.}
Table~\ref{tab:7blog} reproduces the Qwen2.5-7B budget sweep exported
from \texttt{results/processed/7b/summary.csv}. Sample sizes reflect the
logged shard ($N{=}54$ for $B \in \{256, 512\}$ and $N{=}53$ for
$B{=}1024$).

\begin{table}[h]
\centering
\caption{Qwen2.5-7B across budgets under 4-bit inference (processed CSV).}
\label{tab:7blog}
\small
\begin{tabular}{lrrrrr}
\toprule
Method & $B$ & $N$ & Acc.\ (\%) & Avg toks & Prem.\ stop (\%) \\
\midrule
fixed       & 256  & 54 & 44.4 & 255 & 0.0 \\
adaptive    & 256  & 54 & 42.6 & 231 & 7.4 \\
\method{}   & 256  & 54 & 44.4 & 243 & 3.7 \\
\midrule
fixed       & 512  & 54 & 90.7 & 466 & 0.0 \\
adaptive    & 512  & 54 & 79.6 & 286 & 14.8 \\
\method{}   & 512  & 54 & 83.3 & 316 & 11.1 \\
\midrule
fixed       & 1024 & 53 & 90.6 & 813 & 0.0 \\
adaptive    & 1024 & 53 & 79.3 & 305 & 18.9 \\
\method{}   & 1024 & 53 & 83.0 & 336 & 15.1 \\
\bottomrule
\end{tabular}
\end{table}

\paragraph{Qwen2.5-14B partial sweep.}
Table~\ref{tab:14b} reports the 14B sweep with $N \in \{34, 35\}$ items
per setting due to Colab session limits. We list these explicitly to
avoid overstating the sample size.

\begin{table}[h]
\centering
\caption{Qwen2.5-14B across budgets under 4-bit inference (partial $N$).}
\label{tab:14b}
\small
\begin{tabular}{lrrrrr}
\toprule
Method & $B$ & $N$ & Acc.\ (\%) & Avg toks & Savings (\%) \\
\midrule
fixed       & 512  & 35 & 88.6 & 455 & ---  \\
adaptive    & 512  & 35 & 82.9 & 239 & 47.5 \\
\method{}   & 512  & 35 & 85.7 & 269 & 40.8 \\
fixed       & 1024 & 34 & 91.2 & 859 & ---  \\
adaptive    & 1024 & 34 & 82.4 & 235 & 72.6 \\
\method{}   & 1024 & 34 & 85.3 & 266 & 69.0 \\
\bottomrule
\end{tabular}
\end{table}

\paragraph{Reproducibility checklist.}
\begin{itemize}
  \item Greedy decoding (\texttt{do\_sample=False}) with fixed prompt
        templates removes sampling stochasticity; the only remaining
        sources of variability are CUDA non-determinism and shard
        composition, both of which are documented in the repository.
  \item All controller hyperparameters are stored in versioned
        configuration files; tables in this appendix were generated
        directly from the released processed-results CSVs.
  \item Figures and tables can be regenerated from raw and processed logs
        without re-running the model, as long as the released archives
        in \texttt{results/} are present.
  \item A typical T4 wall-clock budget for the full per-model sweep
        (items $\times$ methods $\times$ budgets) is approximately
        $25$ minutes for 3B ($50 \times 3 \times 3 = 450$ calls),
        $80$ minutes for 7B ($54 \times 3 \times 3 = 486$ calls), and
        $110$ minutes for 14B ($35 \times 3 \times 2 = 210$ calls),
        dominated by autoregressive decoding rather than by controller
        overhead. Stronger GPUs (A100 / RTX 4090) typically finish
        $3$--$6\times$ faster.
\end{itemize}

\end{document}